\documentclass[10pt, conference, compsocconf]{IEEEtran}
\ifCLASSINFOpdf
  \usepackage[pdftex]{graphicx}
  \graphicspath{{../pdf/}{../jpeg/}}
  \DeclareGraphicsExtensions{.pdf,.jpeg,.png}
\else
\fi
\begin{document}
%
\title{Novelty Detection via Network Saliency in Visual-based Deep Learning}


\author{\IEEEauthorblockN{Valerie Chen}
\IEEEauthorblockA{
Yale University\\
v.chen@yale.edu}
\and
\IEEEauthorblockN{Man-Ki Yoon}
\IEEEauthorblockA{
Yale University\\
man-ki.yoon@yale.edu}
\and
\IEEEauthorblockN{Zhong Shao}
\IEEEauthorblockA{
Yale University\\
zhong.shao@yale.edu}
}


%


\maketitle

\begin{abstract}
Machine-learning driven safety-critical autonomous systems, such as self-driving cars, must be able to detect situations where its trained model is not able to make a trustworthy prediction. Often viewed as a black-box, it is non-obvious to determine when a model will make a safe decision and when it will make an erroneous, perhaps life-threatening one. Prior work on novelty detection deal with highly structured data and do not translate well to dynamic, real-world situations. This paper proposes a multi-step framework for the detection of novel scenarios in vision-based autonomous systems by leveraging information learned by the trained prediction model and a new image similarity metric. We demonstrate the efficacy of this method through experiments on a real-world driving dataset as well as on our in-house indoor racing environment.
\end{abstract}


%
\IEEEpeerreviewmaketitle

\section{Introduction}

One of the most successful examples at the forefront of autonomous systems is self-driving cars, vehicles with the ability to sense their environment and navigate the road without human direction and supervision. The advent of autonomous vehicles will make the streets safer, reducing the number of accidents by up to 90 percent by removing human-error caused accidents \cite{mckinsey}. The technology currently driving this movement is deep learning. These technologies are powered by machine learning algorithms trained extensively on mass amounts of data collected from driving in real life and in simulation \cite{waymo, bojarski2016end}.

As this technology rapidly progresses, there is an increasing concern with regard to safety. Deep neural networks are not trained with safety concerns in mind and are themselves a cause of worry due to the lack of transparency in its decision-making process. Trust in safety-critical autonomous systems like self-driving cars is tied directly to the amount of knowledge we have of the internal decision-making mechanism. It is non-trivial to determine what types of situations a model is able to make a safe decision and what types it will make an erroneous and perhaps life-threatening one. Recent works have shown that deep learning networks are not robust and simple adversarial attacks such as the addition of noise can drastically change the prediction of the model \cite{kurakin2016adversarial, tian2018deeptest, engstrom2017rotation}. 


This paper proposes a multi-step framework to detect novel scenarios in a vision-based autonomous driving system. We demonstrate the feasibility of quantifying the novelty of an encountered situation ``in the wild'' by applying the approach to two driving datasets, one from real-world driving and one from our in-house racing environment. 

\section{Problem Description}

Given a large set of training images $D$ and a trained deep learning model $M$, we consider a problem of determining whether a new image $d \notin D$ is \emph{novel} and thus would produce an output $M(d)$ that is \emph{not trustworthy}. In other words, we aim to determine if $d$ is similar to the dataset $D$ that the model is trained on. It should be noted that we do not aim to determine if $M(d)$ is correct or not; 
the trained model may output a correct prediction even when the input is dissimilar to $D$. 
In this paper, we target an application of predicting steering angle for a given road image \cite{bojarski2016end}, which is a regression problem. We assume that the model has been trained on sufficient examples of the problem to be able to produce accurate output for input that is similar to ones it has seen in the training phase. 

The difficulty of the problem arises from the high dimensional space from which the images are sampled from, that is the highly diverse driving environment. 
A robust novelty detection method should be able to detect images not only from an entirely unseen environment but also from altered, yet similar images of a seen environment that have been slightly modified through adversarial attacks. We propose a method to make strides on solving these problems. 

\subsection{Related Work}

Previous approaches to novelty detection have been focused on designing and training one-class classifiers. In a one-class classifier, all data points in the training set are considered within the target class and all other possible data points are considered novel, so the novel class is disproportionately large compared to the target class. These type of one-class approaches \cite{ruff2018deep, sabokrou2018adversarially} have been largely focused on classification applications. Experimental results have been performed on datasets including MNIST (handwritten digits), CIFAR10 (10 object classes), and Caltech-256 (256 object categories), which are all highly structured and distinct object datasets.

In the context of robotic systems, work by Richter and Roy \cite{richter2017safe} has provided preliminary results for an approach to novelty detection in a model race car system. They trained the race car to drive through a hallway using a collision avoidance network. To detect a novel environment, they trained a single autoencoder, which is an example of a one-class classifier, on the training images with a loss function defined as the mean square reconstruction error of the image. The authors note that even their environment is still highly structured and not representative of real driving environments. Hence, their method of utilizing an autoencoder to memorize training images produced good results. We find that such a method that extracts features from raw images using a stand-alone autoencoder will not generalize to real-world situations where input images are more complex and contain many irrelevant features (e.g., the shape of clouds or the color of shop signs should not affect the steering prediction of a driving model).  
Thus we propose an approach that leverages salient features from the trained model for novelty detection, which we detail in the next section. 

\section{Technical Detail}

\subsection{Overview of Framework}


\begin{figure}[t]
\centering
  \includegraphics[width=\columnwidth]{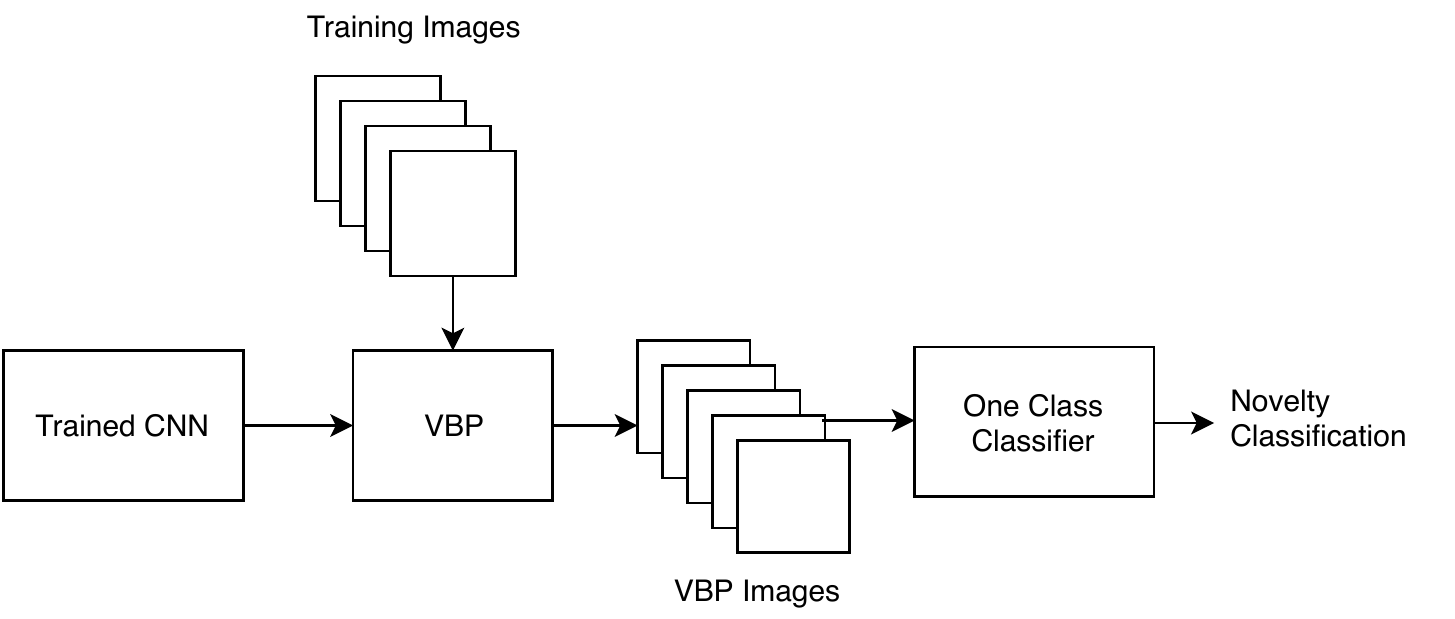}
    \caption{The proposed two-layer framework for novel input detection.}
    
   \label{figure2}   
\end{figure}

Figure~\ref{figure2} presents our approach to novel input detection 
for visual-image based machine learning applications. We assume a CNN (Convolutional Neural Network) model has been trained to map input images to output steering angles. The model we consider is modeled off of the steering angle prediction convolutional network presented in \cite{bojarski2018visualbackprop}. However, our method is not limited to this particular network architecture. The training images we consider will be discussed further in later sections.

Given such a CNN model and associated training images, we apply the following steps: at the preprocessing layer, extract salient features from the images using VisualBackPropogation (VBP) \cite{bojarski2018visualbackprop}, then feed these images to the one-class classifier to learn a representation of the extracted features. In our approach we select an autoencoder as the one-class classifier. We utilize a feedforward autoencoder with 3 hidden fully-connected layers (64, 16, 64 nodes respectively at each layer) with ReLU activation and a sigmoid output layer. We use relatively low resolution (60x160) images so the output layer has dimensions 9600. Each image is first converted to grayscaled and then normalized to range between [0,1]. We divide the training and testing set with an 80/20 split and train with a mini-batch size of 32. The next section presents technical details of the input image preprocessing phase and the autoencoder loss function.

The same framework is used in the testing phase; once we have the both the VBP model and one-class classifier, we preprocess a new image with the VBP of the prediction model. Novel images would likely produce garbled results, and thus be classified as novel, since the network will not be able to find the features it was trained on.

\subsection{Extracting Salient Features using Visual Backpropogation}

\begin{figure}[t]
\centering
  \includegraphics[scale=0.35]{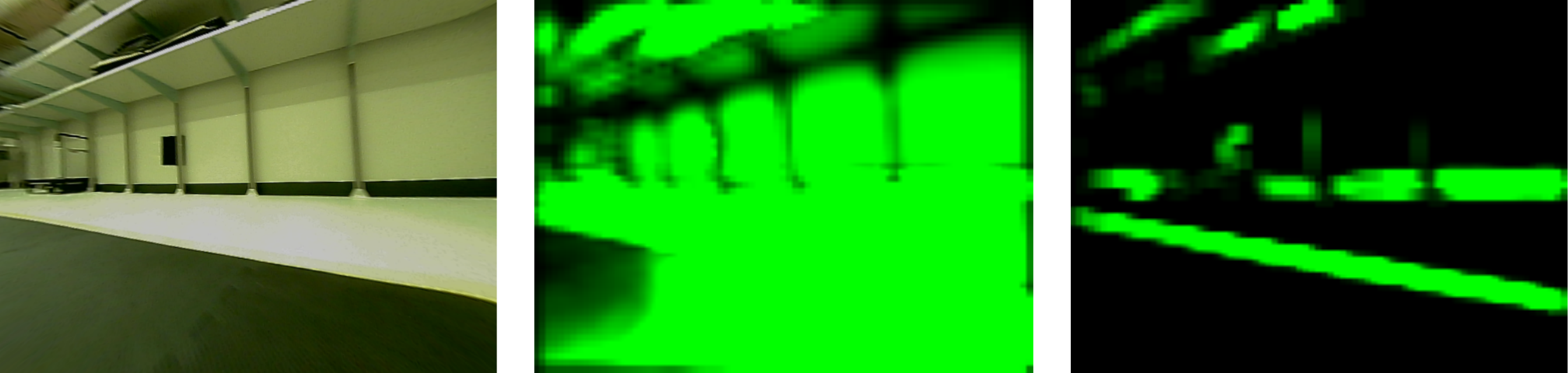}
    \caption{A preliminary experiment on our in-house data to demonstrate that VBP masks are tied to learned features. (Left) Original image view of the road (Middle) Generated VBP mask on network trained with random steering angles (Right) Generated VBP mask on network trained with actual driving angles. This demonstrates that given meaningful input and output, VBP can extract key areas of an image such as the edge of the road.}
    
   \label{figure3}   
\end{figure}

Deep learning models are often viewed as a "black box" system because simple inspection of numerical weights of the network do not convey its decision-making process. Recent developments on network saliency visualization methods like VBP aim to give insight on what aspects of an input caused the output (see Figure~\ref{figure3}). In particular, VBP identifies sets of pixels of the input image that contribute most to the predictions made by a trained CNN through combining feature maps from deeper convolutional layers with more relevant information with higher resolution feature maps of shallow layers. The outputted mask is computed through scaled and averaged deconvolutions of each internal convolution layer after a forward pass.

We propose to utilize VBP as the preprocessing layer to extract the salient features of an input image based on what the trained model learned for making predictions in a noisy, dynamic real-world environment. 
VBP has been demonstrated to be order of magnitude faster than other network saliency visualization methods (such as \cite{bach2015pixel}) that produce comparable, making it an appropriate choice for real-world systems where real-time decision making is required.

\begin{figure*}[t]
\centering
  \includegraphics[width=0.8\textwidth]{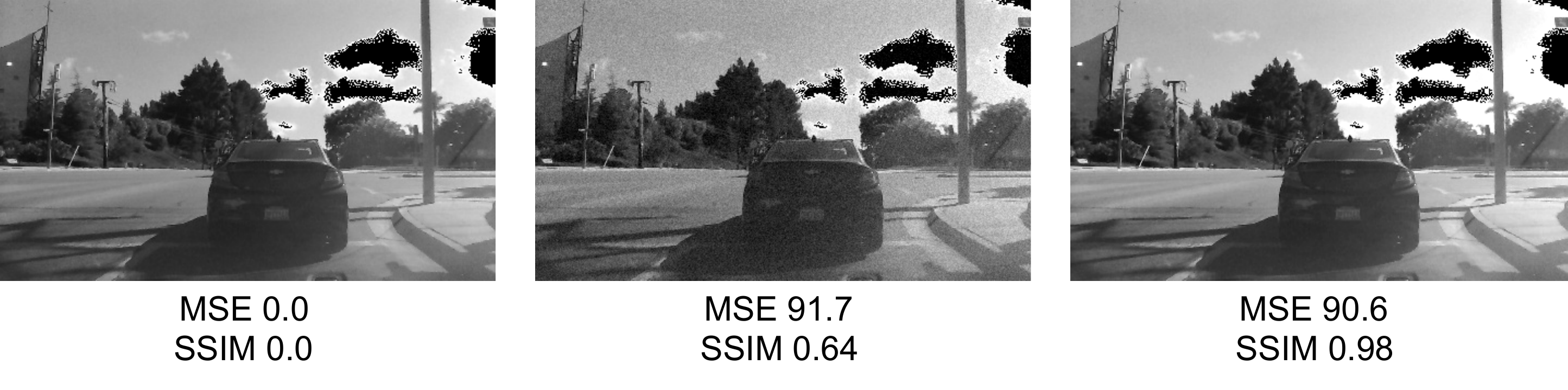}
    \caption{MSE and SSIM of (Left) Original Image (Middle) Added Gaussian Noise (Right) Increased brightness. 
     }
   \label{figure4}   
\end{figure*}

\subsection{Training VBP Images using Structural Similarity Index}

Previous work on novelty detection using autoencoders in \cite{richter2017safe, japkowicz1995novelty} employ pixel-wise mean squared error (MSE) for the metric of image similarity, where the loss function between the image $x$ and its reconstruction $y$ is defined as:
\[MSE(x, y) = \frac{1}{K} \sum_{k=1}^{K}(x[k] -y[k])^2,\]
where $K$ is the number of pixels, $x[k]$ and $y[k]$ are the intensity values of the $k^{th}$ pixel of the image. Both works perform a threshold test on the MSE for the novelty classification; an image is classified as novel if its MSE falls outside of the $99^{th}$ percentile of the empirical CDF (cumulative distribution function) of the distribution of $MSE$ losses in the training set. For visually distinct categories of images, the distributions of reconstruction errors are clearly separable, thus the value of the threshold is not critical. However, in \cite{richter2017safe}, the authors note that this pixel-wise loss method will not be effective for training images that have more variation due to, for example, highly noisy and dynamic environments such as real-world driving conditions. 

Wang et al. \cite{wang2009mean} proposed an alternative metric for image similarity called Structural Similarity Index (SSIM). SSIM takes in two images and reports a similarity score ranging from -1 to 1, in a convolution-type sliding window manner. The details how each perceptual aspect (luminance, contrast, and structure) is derived is presented in \cite{wang2009mean}. Hence, we briefly review the general motivation of each: luminance $I(x,y)$ correlates with average pixel intensity, contrast $C(x,y)$ with the standard deviation, and structure $S(x,y)$ with covariance. These aspects are combined to form the SSIM score, given $x$ and $y$ are 11x11 patches of an input and reconstructed image respectively: 

\[SSIM(x,y) = I(x,y)^\alpha C(x,y)^{\beta}S(x,y)^\gamma,\] 
where we set $\alpha = \beta = \gamma = 1$ and reduce to this form:

\[SSIM(x,y) = \frac{(2\mu_x \mu_y + c_1)(2\sigma_{xy}+c_2)}{(\mu_x^2+\mu_y^2+c_1)(\sigma_x^2+\sigma_y^2+c_2)},\]
where $\mu_x$ (resp. $\mu_y$) is the average pixel intensity values of $x$ (resp. $y$), $\sigma_x^2$ (resp. $\sigma_y^2$) is the variance of $x$ (resp. $y$), $\sigma_{xy}$ is the covariance of $x$ and $y$, and $c_1$ and $c_2$ are smoothing constants to prevent division by 0.

SSIM has been applied as a loss function in training image reconstruction autoencoders \cite{snell2017learning}, where they demonstrated that traditional $L_2$ MSE loss functions produce blurry results whereas SSIM is grounded in human perceptual judgments. 

A key benefit of SSIM over MSE is that SSIM is normalized such that value always fall within the specified range where 1.0 means perfect correspondence, 0.0 means no correspondence, and -1.0 means perfect negative correspondence between the two images, thus intermediate values have relative significance.\footnote{When optimizing for MSE loss, we desire smaller values as a MSE of 0.0 means perfect reconstruction. On the other hand, when optimizing for SSIM loss, we want to maximize the SSIM value as a 1.0 means perfect reconstruction.}  For MSE, however, 
the range of values is highly dependent on the range of intensities and distribution of pixel values, which is normalized out in the SSIM metric. Therefore, it is difficult to compare reconstructed outcomes given two MSE values of two different images. 

Figure \ref{figure4} illustrates the difference between SSIM and MSE through an example. While both modified images (middle and right) have been engineered to result in similar MSE purely based on pixel-wise loss, the one with Gaussian noise added has a significantly lower SSIM than the one with brightness changed. This aligns with findings that convolutional network tend to be more robust to brightness changes than noise \cite{dodge2016understanding}, so there should in fact be a greater decrease in similarity for noise than brightness.

\begin{figure}[h!]
\centering
  \includegraphics[width=\columnwidth]{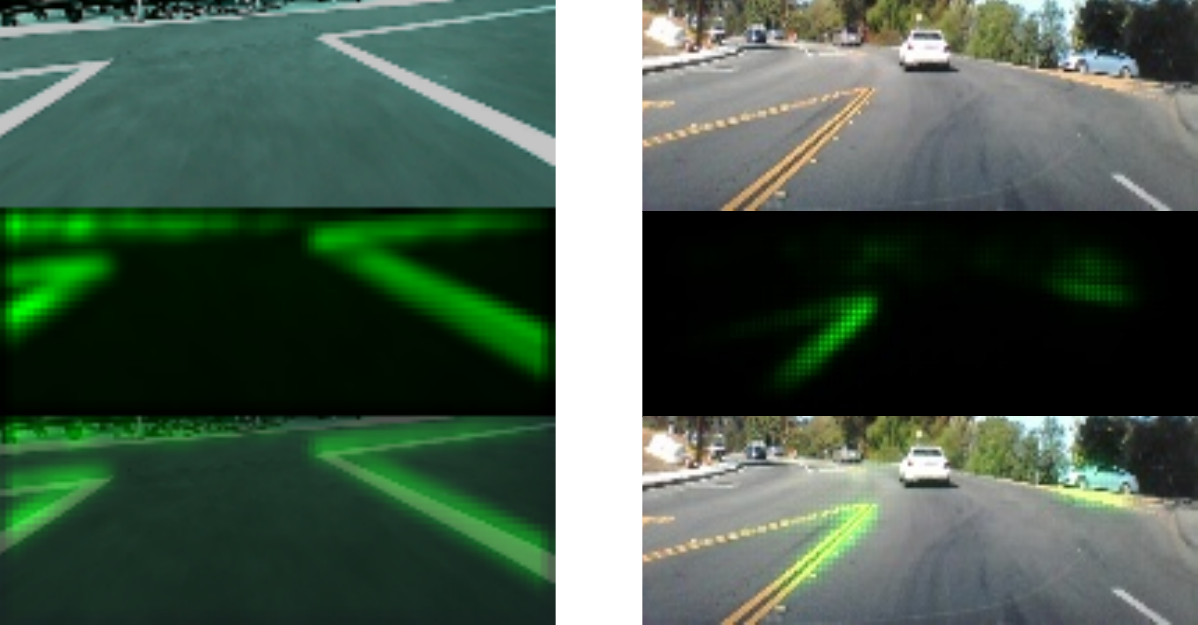}
    \caption{Example VBP output for a road image for our in-house driving data on left and Udacity data on right. (Top: input image, middle: VBP mask, bottom: mask overlaid on input image to demonstrate reasonable activations as a human driver would expect.)}
   \label{figure5}   
\end{figure}
\section{Evaluation}

\subsection{Datasets}

\newcommand\udacity{$\mathbf{DS_U}$}
\newcommand\ydriving{$\mathbf{DS_I}$}

\begin{figure*}[t]
\centering
  \includegraphics[width=0.9\textwidth]{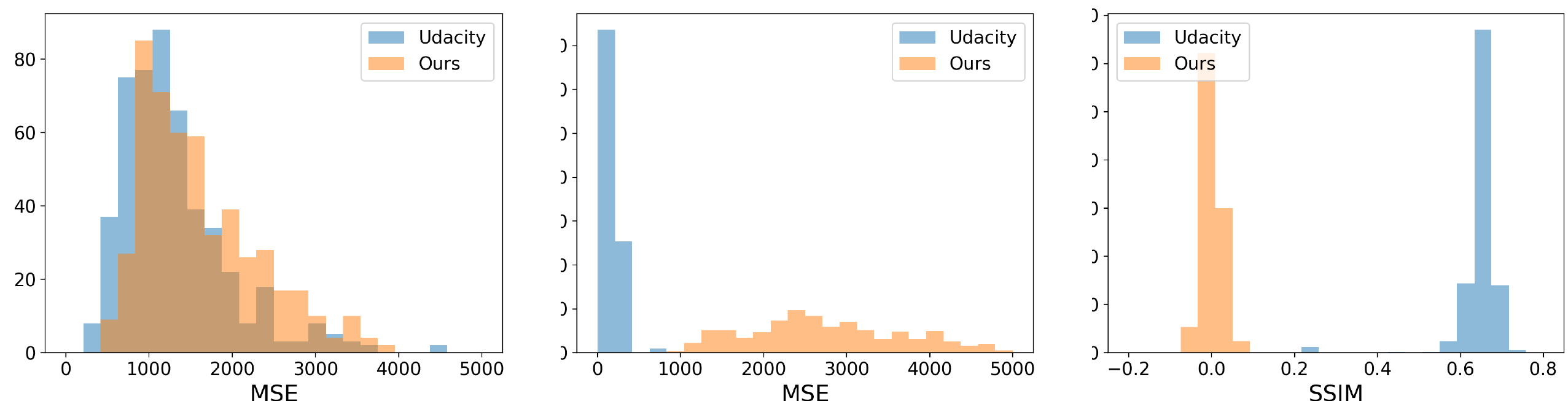}
    \caption{Histogram comparison for original images with MSE loss (Left), VBP images with MSE loss (Middle), and VBP images with SSIM loss (Right) when the network is trained on Udacity images, \udacity{}.
    }
   \label{figure7}   
\end{figure*}

For our experiments, we primarily work with the Udacity self-driving car dataset \cite{udacitydataset}, which consists of over 45,000 images collected from actual driving in Mountain View, CA. The dataset has annotated images with associated steering angles. We also collected an in-house dataset from a model car driving in an indoor racing environment. The roads in our model self-driving environment have varied surroundings and backgrounds, which provides for more variety than environments and structured datasets studied in \cite{richter2017safe}. Hereafter, we denote the Udacity dataset and our in-house dataset by \udacity{} and \ydriving{}, respectively. 

In \cite{bojarski2018visualbackprop}, the authors demonstrate that VBP produces a qualitatively reasonable visualization for \udacity{} when trained to predict steering angles. Figure \ref{figure5} presents example VBP masks overlaid on the input image for both datasets \ydriving{} and \udacity{}.

\subsection{Experiments}

Given the two datasets \udacity{} and \ydriving{}, we demonstrate through a set of experiments the utility of the proposed novel input detection approach. We utilize each dataset as the target class and the other as the novel class to demonstrate capability to distinguish between two different driving datasets. As previously noted, the network is trained to predict the steering angle given the current image view of the road. We also explore the effect of adding noise to the images.

\vspace{0.5\baselineskip}

\subsubsection{MSE vs. SSIM}

\begin{figure}[b]
\centering
  \includegraphics[scale=0.7]{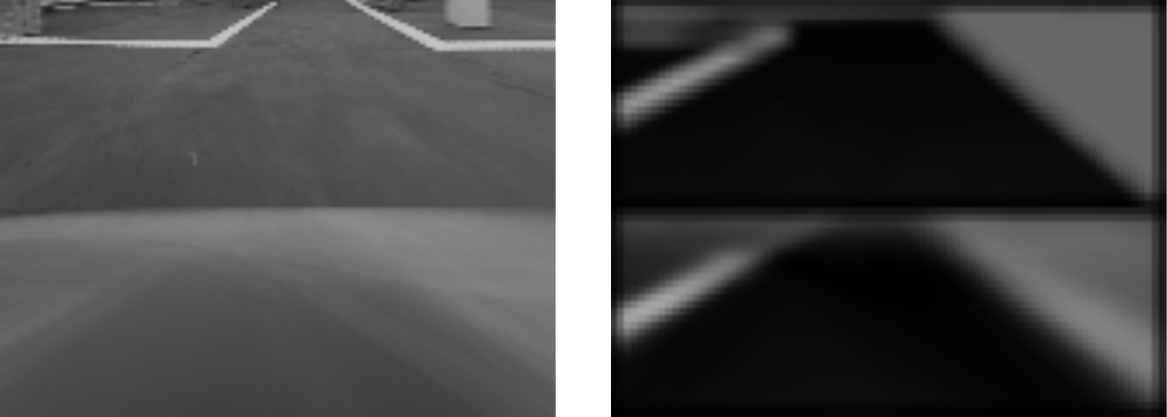}
    \caption{(Left) Image reconstruction of an original image with MSE loss (Right) Image reconstruction of a VBP image with SSIM loss. On top is the input image and bottom is the reconstructed image by our autoencoder.}
   \label{figure6}   
\end{figure}

Figure \ref{figure6} compares our proposed approach utilizing VBP images trained with SSIM loss against the prior approach in \cite{richter2017safe} utilizing the original images with MSE loss. The latter produces a blurry reconstruction even for a target class image (i.e., not novel) causing it to be not visually distinguishable from the blurry reconstruction of a novel class image. Therefore, the autoencoder with MSE loss of \cite{richter2017safe} is prone to fail to separate out novel images when given data with more variation.

Our results of comparing the two loss functions confirm results from \cite{snell2017learning} where SSIM produces reconstructions that are more structurally similar to the input image. We find that with SSIM, we are able to clearly distinguish the target class data, which have clean reconstructions, from novel data, which have blurry reconstructions by the autoencoder. We support these findings with histogram loss comparisons in the next two sections. 

\vspace{0.5\baselineskip}

\subsubsection{Dataset Comparison}

\begin{figure*}[t]
\centering
  \includegraphics[width=0.65\textwidth]{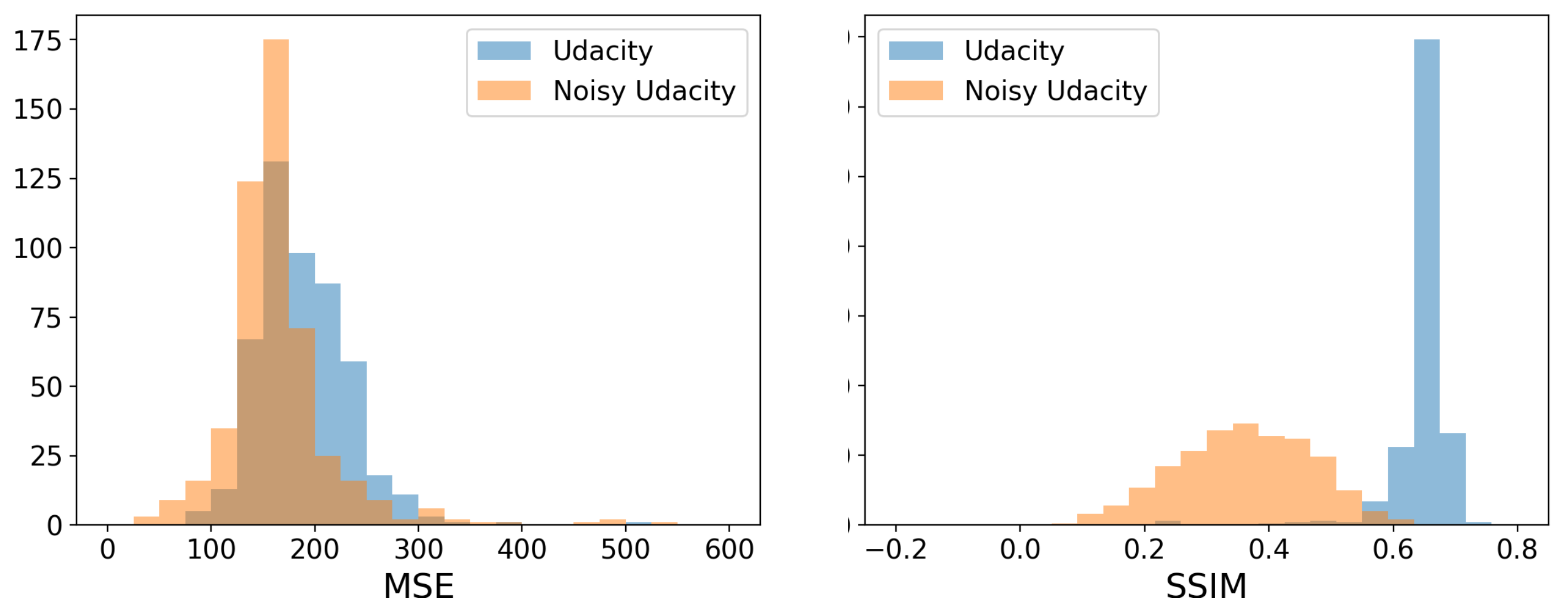}
    \caption{Histogram comparison for MSE (Left) and SSIM (Right) values of \udacity{} VBP images versus noisy \udacity{} VBP images. An MSE loss is not able to distinguish noisy images while SSIM is able to separate the two distributions. We also performed the same experiment on MSE loss on original images and found similar results as the left histogram plot above.
    }
   \label{figure8}   
\end{figure*}

We compare our propose approach with the prior method \cite{richter2017safe} that utilizes a stand-alone autoencoder with MSE loss. In our approach, we trained the autoencoder on VBP images of \udacity{}. Similarly we used the VBP images of \ydriving{} as the novel data. For training, we used $80\%$ of \udacity{}. For testing, we randomly sampled 500 images from the rest. We also randomly sampled 500 images from \ydriving{} for testing. The right plot of Figure \ref{figure7} shows that we were able to achieve an average SSIM value of about 0.7 out of 1.0 for the testing \udacity{} images while \ydriving{} images had almost 0 similarity with its reconstructions. The method is able to clearly distinguish \ydriving{} from \udacity{} as all of \ydriving{} testing samples were classified as novel. We performed the same comparison using \udacity{} as the training set and \ydriving{} as the novel set with MSE loss on original images (the left plot) as well as MSE loss on VBP images (the middle plot). MSE loss on VBP images improves upon MSE loss on original images, while SSIM loss on VBP images most clearly separates the two class distributions. This demonstrates that VBP is necessary to filter out unnecessary details in the original training images and results further improve with using SSIM loss.

\vspace{0.5\baselineskip}

\subsubsection{Noise Detection}

For this experiment, the novel dataset was created by adding Gaussian noise to a sampled set of the \udacity{} training set using the same method as above. Then the images were passed through VBP to generate the network salience map of the noisy image. As expected, the VBP images of the noisy images were also garbled looking and also visibly noisy compared to its original counterpart. Figure \ref{figure8} shows both the MSE and SSIM distributions of the original and noisy images. The separation between noisy data and original data is smaller in both cases as some key features, i.e. lane markings, from the original data can still be found in the noisy images whereas the separation from data sampled from a different dataset is much greater as shown in Figure~\ref{figure7}. In line with findings from \cite{snell2017learning}, we also observed that SSIM is superior over MSE when differentiating finer grain detail, i.e. noise in this case.

Due to space limitations, we do not present results for training on \ydriving{ }and using \udacity{} as novel data, but we were able to find comparable results. We note that \udacity{} is a more varied dataset compared to our \ydriving{}, which means these results are more difficult to achieve on the less structured dataset.

\section{Conclusion}

In this paper, we proposed a multi-step framework to detect input data that is novel with respect to the training data that the network model was trained on.  Our approach employs an autoencoder based method with visual saliency preprocessing and a novel loss function. We apply our method to visual navigation for self-driving car to demonstrate on multiple experiments that our proposed method outperforms prior work and the benefits of each key component: the preprocessing phase and loss function in the framework. The proposed approach makes strides towards building a safer machine learning based autonomous systems where trust in the model is incredibly important. 


\section*{Acknowledgment}

This work is supported in part by NSF grants 1521523, 1715154, and 1763399. Any opinions, findings, and conclusions or recommendations expressed here are those of the authors and do not necessarily reflect the views of sponsors.



\bibliographystyle{IEEEtran}
%

\bibliography{refs}

\end{document}